\documentclass[conference]{IEEEtran}
\IEEEoverridecommandlockouts
\usepackage{cite}
\usepackage{amsmath,amssymb,amsfonts}
\usepackage{algorithmic}
\usepackage{graphicx}
\usepackage{textcomp}
\usepackage{comment}
\usepackage{xcolor}
\usepackage{subfigure}
\def\BibTeX{{\rm B\kern-.05em{\sc i\kern-.025em b}\kern-.08em
    T\kern-.1667em\lower.7ex\hbox{E}\kern-.125emX}}
\DeclareMathOperator*{\argmin}{arg\,min}
\begin{document}
\title{TACO: Rethinking Semantic Communications with Task Adaptation and Context Embedding}
\author{\IEEEauthorblockN{Achintha Wijesinghe$^*$, Weiwei Wang$^*$, Suchinthaka Wanninayaka$^*$, Songyang Zhang$^\dagger$, Zhi Ding$^*$ }
\IEEEauthorblockA{$^*$University of California at Davis,  Davis, CA, USA, 95616\\$^\dagger$University of Louisiana at Lafayette, Lafayette, LA, USA, 70504 }}


\maketitle
\begin{abstract}

Recent advancements in generative artificial intelligence
have introduced groundbreaking approaches to innovating next-generation semantic communication,
which prioritizes conveying the \textit{meaning} of a message rather than merely transmitting raw data.
A fundamental challenge in semantic communication lies in accurately identifying and extracting the most critical semantic information while adapting to downstream tasks without degrading performance, particularly when the objective at the receiver may evolve over time. 
To enable flexible adaptation to multiple tasks at the receiver, this work introduces a novel semantic communication framework, which is capable of jointly capturing task-specific information to enhance downstream task performance and contextual information. Through rigorous experiments on popular image datasets and computer vision tasks, our framework shows promising improvement compared to existing work, including superior performance in downstream tasks, better generalizability, ultra-high bandwidth efficiency, and low reconstruction latency.

\end{abstract}


\section{Introduction}
Next-generation communication systems are expected to support the surge in data-intensive applications with the increasing demand to
handle a copious amount of multimodal data generated from intelligent devices, including those from smart sensors, ecosystems of the Internet of Things, mixed reality, and autonomous vehicles~\cite{mm1}. 
To enable wireless communications with the capacity to satisfy the request from the receiver end with ultra-high bandwidth efficiency in the big data era, semantic communication (SemCOM) has emerged as a transformative paradigm, which shifts data transmission from faithful bitwise recovery of source data to conveying its most critical semantic \textit{meaning}~\cite{diffgo}. 

Existing semantic communications originate from the idea of learning for communications, where deep neural networks (DNN) are trained to embed the original data into semantic representations at the transmitter, which are transmitted to the receiver end for learning-based recovery. Classic frameworks include end-to-end DNN frameworks~\cite{9398576,9322296,10038754}. However, these conventional end-to-end SemCom frameworks lack interpretability of semantic meaning and suffer from limited bandwidth improvement. With the development of artificial intelligence (AI), SemCOM based on generative AI models (Gen-AI) has attracted significant attention. One category of forefront approaches relies on the diffusion models due to their power in data regeneration and flexible multimodal conditioning \cite{gesco}. In diffusion-based SemCom, semantic representations, such as segmentation maps and image captions, are extracted at the transmitter, while a diffusion model is
deployed at the receiver for data regeneration conditioned on the transmitted semantic representation \cite{diffgon}. Despite some successes, existing diffusion-based methods rely on pre-selected semantic representations with human knowledge, limiting their capacity for automation and generalization. On the other hand, the randomness introduced in the sampling stage may lead to unexpected artifacts. Although some work ~\cite{diffgo,diffgo+} attempts to address randomness using feedback control, the reconstruction time can sometimes be impractical. To handle the computational burden, recent interests have focused on incorporating representation learning with generative AI. One potential solution is based on Vector Quantized Variational Autoencoder (VQ-VAE)~\cite{vqvae}, which explores the semantic latent space~\cite{lamigo}. Other examples also include VQ-based token communications, such as those for cross-modality~\cite {mahadi1} and multi-access communications~\cite{mahadi2}. Although these methods can reduce the computational complexity, they still focus on optimizing the performance for predefined tasks~\cite{multi1}, which cannot efficiently address dynamic tasks, particularly suffering from heavy retraining of the neural networks.

To address the aforementioned challenges, we introduce a novel semantic communication framework, namely \textbf{TACO}, considering both \textbf{T}ask-\textbf{A}daptive information and data \textbf{CO}ntext. Taking vision tasks as an example, we embed a given image into either context representation or task-specific information, facilitating the SemCom system with promising performance on predefined tasks while preserving the potential to address new goals. Specifically, we design an embedding space for both context and goal-oriented information by utilizing the vector quantization of a pretrained VQ-VAE~\cite{vqvae} network, where GradCAM~\cite{gradcam} is equipped to understand the region of interest for the downstream task model. Offering a convenient fusion of context and task-dependent information at a latent level, our proposed framework can significantly reduce the communication and computation cost, with enhanced flexibility for task adaptation.

Our main contributions can be summarized as follows:
\begin{itemize} 
\item We design a novel SemCom framework capable of optimizing performances for predefined tasks while being equipped with flexibility for changing receiver goals. We design a novel scheme to identify two essential semantic representations, i.e., context and task-driven information, for a robust semantic communication system. 

\item We propose an innovative latent-level communication and fusion mechanism using a lightweight VQ-VAE model, which significantly reduces the data reconstruction latency compared to existing diffusion-based approaches.

\item We introduce a local semantic feedback mechanism at the transmitter, enabling the selection of the most informative and compact representations to enhance both bandwidth efficiency and downstream task performance.

\item We validate the effectiveness of our approach through extensive experiments on several widely used image datasets in vision tasks compared to state-of-the-art (SOTA) SemCom frameworks.
\end{itemize}

\vspace{-0.2cm}
\section{Preliminary}
\vspace{-0.1cm}
In this section, we first introduce the preliminaries of VQ-VAE and provide an alternative formulation of SemCom.
\vspace{-0.1cm}
\subsection{VQ-VAE}
VQ-VAE~\cite{vqvae} is composed of an encoder $E(\cdot)$ to embeds data as a latent representation by learning important features, and a decoder $D(\cdot)$ to recover the original image from this latent representation. For this, nearest neighbor-based vector quantization is used with the following training objective 
\begin{multline}
    \mathcal{L}_{VQ} = ||\hat{\mathbf{x}} - \mathbf{x}||^2_2  + ||SG[E(\mathbf{x})] - \mathbf{c}||^2_2 \\
    + \gamma ||E(\mathbf{x}) - SG[\mathbf{c}]||^2_2,
\end{multline}
where $\mathbf{x}$ is the original data,  $\hat{\mathbf{x}}$ is the reconstructed data, $\mathbf{c}$ is the learned codebook vectors, and $SG$ represents the stop gradient operation. Then, quantization $Q(\cdot)$ is introduced for each component $\mathbf{e}_i \in E(\mathbf{x})$, by finding the closed codeword $\mathbf{c}_{k}$ as
\(
k=\argmin_j||\mathbf{e}_k-\mathbf{c}_j||_2^2.
\)

Due to codebook representation, the VQ latent of a given image can be represented using a set of integers. Therefore, the reconstruction image is given by, $\mathbf{\hat{x}}=D(Q(E(\mathbf{x})))$. For brevity, in the following sections, VQ encoder $E(\cdot)$ is assumed $Q(E(\cdot))$, unless otherwise mentioned.

\subsection{Problem Formulation}

We first introduce the conventional formulation of semantic communications.
Suppose that $\mathbf{x}$ represents the information at the transmitter that needs to be communicated to the receiver. In this work, we take images as an example, i.e., $\mathbf{x} \in \mathbb{R}^{C \times H \times W}$ where $C$, $H$, and $W$ denote the number of channels, height, and width of the image, respectively. Let $\hat{\mathbf{x}} \in \mathbb{R}^{C \times H \times W}$ denote the reconstructed message at the receiver.
Given a semantic distortion function $d(\cdot, \cdot)$ and a maximum tolerable semantic degradation $\tau$, our objective is to minimize the communication rate $\mathcal{R}$ over the channel, subject to a semantic fidelity constraint \begin{align} \label{eq:r1} \min \mathcal{R} \quad \text{such that} \quad d(\mathbf{x}, \hat{\mathbf{x}}) \leq \tau. \end{align}
Note that $\hat{\mathbf{x}}$ is unnecessarily to be in the format of original data but a representation containing the key semantic information. 

In practice, the threshold $\tau$ reflects the requirement of the quality of service (QoS) at the receiver. The choice of distortion metric given by $d(\cdot, \cdot)$ can be adapted to the context. For task-specific scenarios, $d(\cdot, \cdot)$ may be defined based on downstream task performance, whereas for more general cases, perceptual metrics such as the Learned Perceptual Image Patch Similarity (LPIPS)\cite{lpips} and the Fréchet Inception Distance (FID)\cite{fid} are commonly used. 

Although solving Eq.~\eqref{eq:r1} optimally is intractable in practice, some direct deep learning approaches attempt to jointly optimize it using a Lagrange multiplier $\lambda$, i.e.,
\begin{align} \label{eq:r2} \mathcal{L} = d(\mathbf{x}, \hat{\mathbf{x}}) + \lambda \mathcal{R} \end{align}

Despite some successes, joint end-to-end training of a deep learning based SemCOM model via Eq. \eqref{eq:r2} could reduce model generalizability. For example, if the model is trained to optimize a classification task, the loss based on a predefined downstream task may limit its flexibility to perform other tasks, such as object detection. An efficient semantic communication shall support the receiver's ability to switch between predefined and new tasks.



In this work, we propose that a semantic communication system should transmit both the context of the message and additional task-dependent information. The task-dependent information enables high performance on predefined tasks, while the context provides a global understanding of the message, allowing adaptability on the receiver’s side. Let $\mathcal{R}_{c}$, $\mathcal{R}$, and $\mathcal{R}_{i}$ denote the transmission rates of the context representation, goal-oriented (GO) message information representation, and image representation, respectively, in some latent space $S$. We first optimize the SemCom system constrained by the following rate inequality for predefined tasks: 
\begin{align} \label{eq:r3} \mathcal{R}_{c} \leq \mathcal{R} \leq \mathcal{R}_{i} + \mathcal{R}_{c} \end{align}

That is, the choice of information level is fixed at deployment. For example, fixed 10\% of task-dependent information is shared with context in the first round of communication and continues to share more task-dependent information upon the request of the receiver. Note that the urge for more information on the receiver side is determined by the confidence level of the predefined task. 
For the above system, the communication rate is bounded below by the context rate and above by the sum of the image and context rates, which
account for possible multi-round communication.
The upper bound on this ``task-specific information" corresponds to the image rate. Hence, the left side of the inequality reflects the minimum information required to convey the context. However, the upper bound in Eq.~\ref{eq:r3} is not tight and can be considered redundant and inefficient due to the fixed nature of the system. Therefore, in our design, we propose a \textit{local semantic feedback} to tighten the upper bound by $\mathcal{R}_{fb}$, where an adaptive information selector is designed following
\begin{align} \label{eq:r4} \mathcal{R}_{c} \leq \mathcal{R} \leq \mathcal{R}_{fb} \leq \mathcal{R}_{i} \leq \mathcal{R}_{i} + \mathcal{R}_{c}. \end{align}

\begin{figure*}[t]
\centerline{\includegraphics[width=1.5\columnwidth]{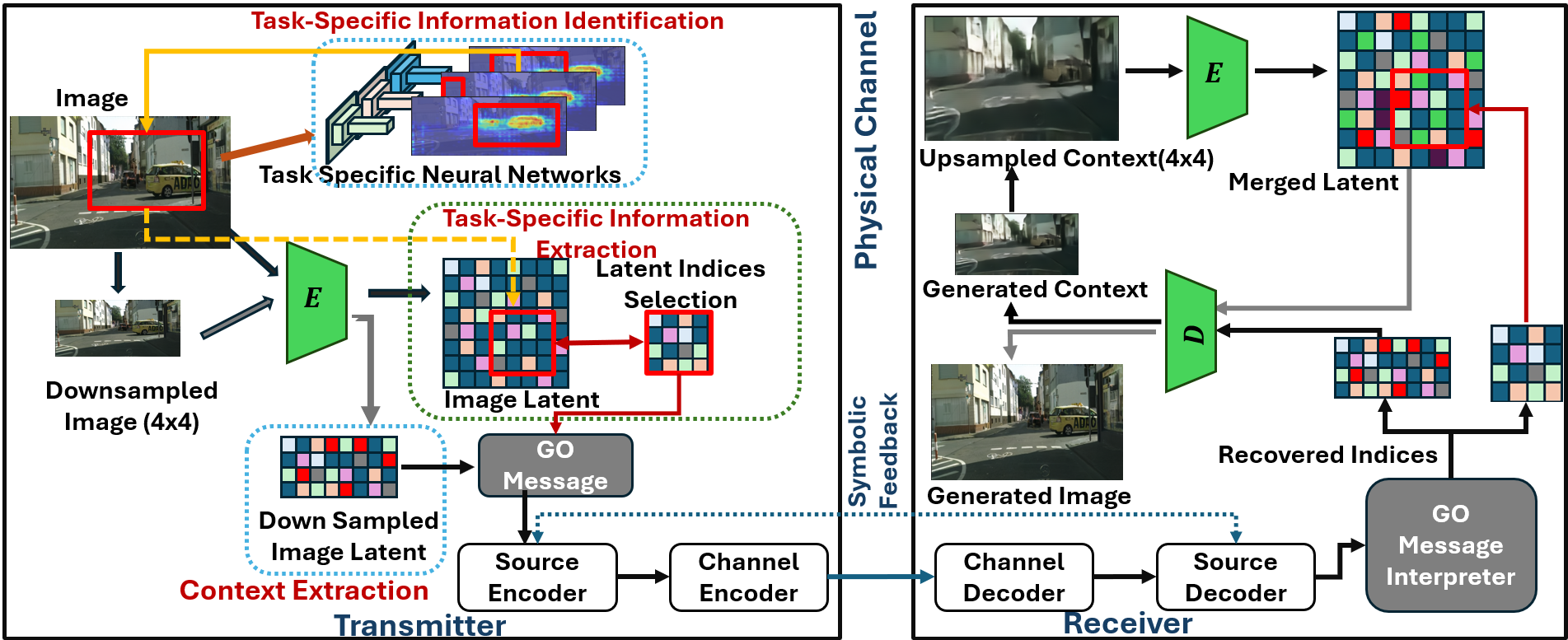}}
\caption{The basic framework of TACO. The transmitter uses the encoder ($E$) to map the image to its latent space. Subsequently, the transmitter downsamples the image by a factor of 4 and utilizes the same $E$ to map the low-resolution image to the same latent space, which serves as the context representation. Meanwhile, the transmitter identifies the task-specific information using a neural network in the pixel domain. Then, the corresponding latent representation is extracted from the image latent. 
At the receiver side, the received context latent is first decoded to the pixel domain and is upsampled by a factor of 4 before being encoded back to the latent space. After deriving the latent, the receiver replaces the corresponding latent embeddings with the received latent indices representing task-oriented information. Finally, the same decoder is utilized to project the merged latent back to the pixel domain.}
\vspace{-0.3cm}
\label{fig:1}
\end{figure*}

\vspace{-3mm}
\section{Method}
We now introduce the details of the proposed TACO SemCom, with overall architecture illustrated in Fig. \ref{fig:1}.

\subsection{Design of the Transmitter}

Our goal in designing the transmitter is to achieve a low transmission rate while maintaining strong performance on downstream tasks. Accordingly, our design choices prioritize low-complexity and compact latent models suitable for real-time communication systems. 
In our transmitter design, we optimize the communication rate $R$ by both representations of context and task-dependent semantic information. 
We utilize the same VQ encoder $E(\cdot)$ to derive both contextual representation and image latent. This step optimizes the memory usage on both sides, allowing an efficient implementation. Detailed modules are introduced as follows.


\subsubsection{Generation of the Image Context}
    Our approach for context $(\zeta)$ generation is inspired by the classical image coding such as JPEG. Considering an image $\mathbf{x} \in \mathbb{R}^{C\times H \times W}$, we downsample it by a factor of $f$ to achieve a downsampled image: $\mathbf{x}_d \in \mathbb{R}^{C\times h \times w}$, where $h = H/f$ and $w = W/f$. As presented in Figure~\ref{fig:1}, we then feed $\mathbf{x}_d$ to the encoder to obtain its latent representation. Now $\mathbf{x}$ has undergone a compression factor of $f^2$ from one side, obtaining a downsampled image latent $\mathbf{z}_d$, which controls $\mathcal{R}_c$ and carries the context of $\mathbf{x}$.

\subsubsection{Identification of the Important Semantic Information}

The central component of our design lies in the identification of salient information within a given image. We propose a mechanism for information identification tailored to predefined downstream tasks. The underlying objective is to discern which pixels the downstream model attends to most significantly. To achieve this, we employ GradCAM~\cite{gradcam} at the transmitter side to analyze the original image and highlight the most informative pixel regions. Subsequently, we project the image $\mathbf{x}$ through the encoder to obtain its latent representation within the VQ domain. Given that our encoder architecture is based on convolutional neural networks, there exists a one-to-one spatial correspondence between the image downsampled by a factor of four and the latent representation $\mathbf{z} = E(\mathbf{x})$. We leverage this correspondence to map the identified salient pixel regions to their respective latent embeddings.

\subsubsection{Backbone System}
In our baseline implementation, we consider various predefined fixed percentages from the set $\{10, 20, 30, 50, 70\}$ to be transmitted as task-specific information along with the context. However, a limitation of this method is the potential requirement for multiple rounds of communication between the transmitter and receiver to fully exchange the identified important content along with the context representation. Hence, shared information in many cases is not optimal and above the required information level. 
Since we transmit both the contextual information and, in the worst case, the latent representation of the entire image, the effective upper bound becomes $\mathcal{R}_i + \mathcal{R}_c$, which is suboptimal. To address this issue, we propose the incorporation of a local semantic feedback mechanism on the transmitter side.
\begin{figure*}[t]
\centerline{\includegraphics[width=1.2\columnwidth]{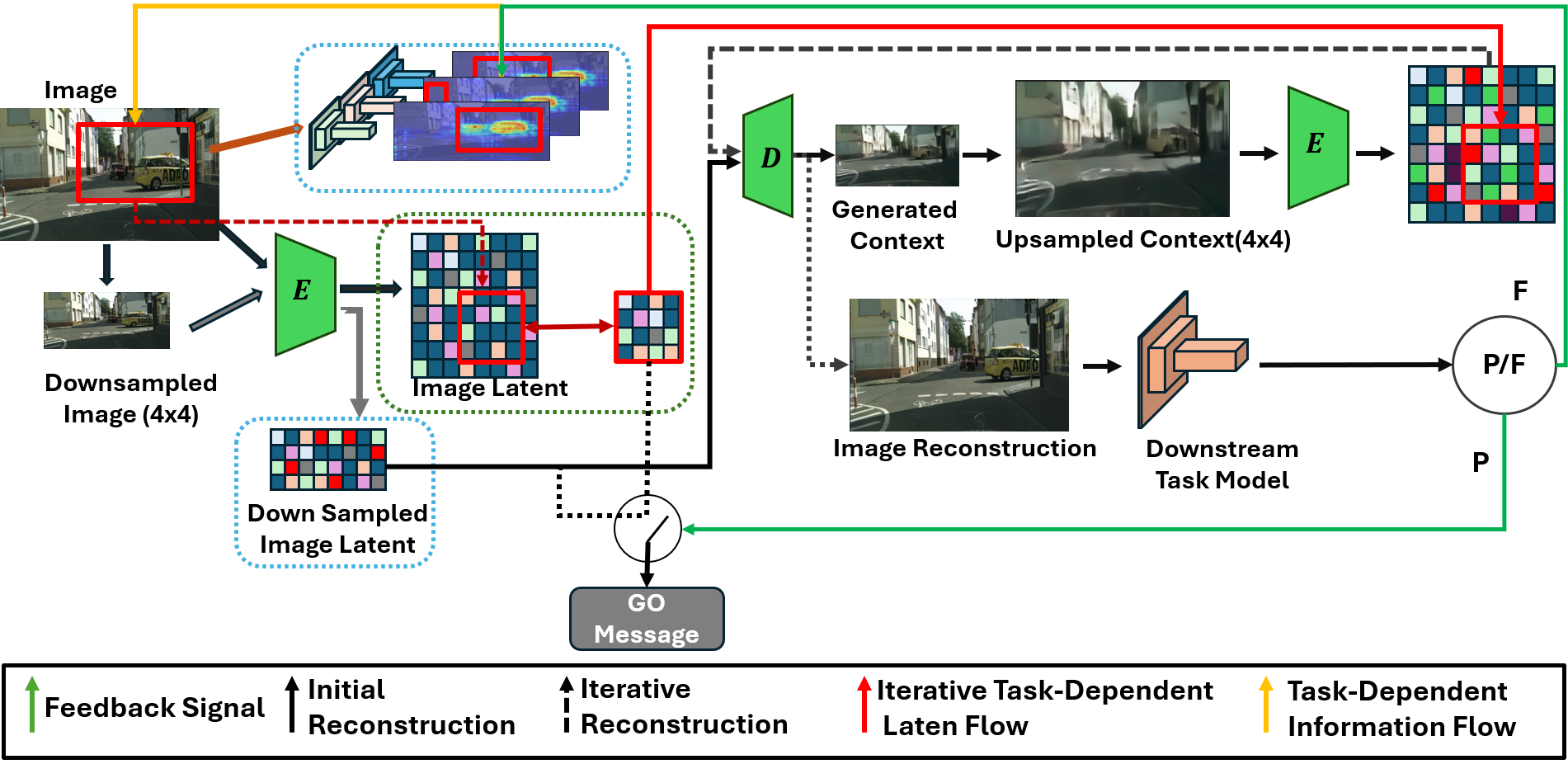}}
\caption{Local feedback for pre-defined tasks. The task-specific information extraction is one-time, but the information is presented partially as different percentages following [10, 20, 50, 100] percentages. }
\vspace{-0.5cm}
\label{fig:2}
\end{figure*}
\subsubsection{Local Semantic Feedback (LSF) }
As illustrated in Figure~\ref{fig:2}, we now introduce an LSF module to further improve the system. 
Assume $\mathbf{x}_u$ to be the upsampled context image to match the original image resolution. i.e. $\mathbf{x}_u = U(D(E(\mathbf{x}_d)))$ where $U(\cdot)$ represent the conventional upsampling using bi-cubic interpolation. If $x_u$ is mapped back to the latent space, i.e., $\mathbf{z}_u = E(\mathbf{x}_u)$, we can identify the corresponding latent embeddings that we are sharing as task-specific information. In the previous steps, we shared both important semantic embeddings and the entire context latent. We exploit this redundancy and share context latent embeddings that are unrelated to predefined semantic information generation.

Let
\begin{equation}
\label{eqn:aprt0}
\mathbf{M}  =
     \begin{cases}
     1 & \text{if coordinate pair $(a,b)\in I_s$ }\\
     0 & \text{else,}\\
    \end{cases}  
\end{equation}
where $I_s$ is the set of coordinates of the positions of the selected latent embeddings of the task-specific information. The latent recovery on the receiver side can be obtained by,
\begin{align}
\label{eqn:aprt1}
    \mathbf{z}_r = (\mathbf{1}_{h\times w} - \mathbf{M})\odot \mathbf{z}_u +  \mathbf{M} \odot \mathbf{z}_i
\end{align}
where $\odot$ is the Hadamard product and 
\begin{equation}
\label{eqn:aprt2}
\mathbf{z}_i  =
     \begin{cases}
     \mathbf{z}(a,b)& \text{if coordinate pair $(a,b)\in I_s$ }\\
     0 & \text{else}\\
    \end{cases}  .
\end{equation}

Therefore, the upperbound rate now reduces to $\mathcal{R}_i$. Now, we can elaborate on our design of LSF on the transmitter side. We first set a task-specific information percentage search set as $\{10, 20, 30, 50, 70, 90, 100\}$. Then, for each percentage in the set, we perform Eq.~\eqref{eqn:aprt0} to Eq.~\eqref{eqn:aprt2} at the transmitter and evaluate the downstream task performance. If the performance is compatible with the ground truth we share the top $p\%$ task-dependent latent embeddings with the context latent embeddings. Otherwise, we simply share the context representation with the other side to facilitate the receiver's other evaluations on the message. Note that, if the sharing information rate $\mathcal{R}$ is higher than the rate of image representation $\mathcal{R}_i$ (i.e., $\mathcal{R}>\mathcal{R}_i$), we simply share the image representation.


\subsection{Design of the Receiver}
At the receiver, the GO message interpreter decodes the context latent representation $\mathbf{z}_c$, the task-specific information representation $\mathbf{z}_i$, and the mask $\mathbf{M}$, following, Eq.~\eqref {eqn:aprt2} for message reconstruction. The first step of message reconstruction is context generation, where we utilize the VQ decoder $D(\cdot)$. Here, $D(\cdot)$ is input resolution agnostic. In order to generate the context, we first feed the context latent representation to the $D(\cdot)$ and upsample it using classical upsampling techniques such as bi-cubic interpolation. Subsequently, the latent representation $\mathbf{z}_u$ is obtained by encoding upsampled images as 
\vspace{-3mm}
\begin{align}
    \mathbf{z}_u = E\left(U\left(D\left(\mathbf{z}_c\right)\right)\right)
\end{align}
Next, the fused latent is obtained using the latent mixup as Eq.~\eqref{eqn:aprt1}. Finally, the image reconstruction is obtained utilizing the same VQ decoder, i.e.,
 \(
    \mathbf{\hat{x}} = D\left(\mathbf{z}_r\right).
\) 

\subsection{Adaptation to New Downstream Tasks}


To address changing downstream tasks, in the first round of communication, we provide context and task-specific information to the receiver. Now, assume a situation where the receiver is interested in performing a new task. For example, the receiver changes from predefined classification to object detection. As a solution,
we feed $\mathbf{\hat{x}}$ to the new downstream task model and evaluate the performance. If the performance needs to be improved, the receiver requests the updated latent representation from the transmitter, which can be done without costing very bandwidth, as the receiver only needs to share the coordinates of the region of interest. 


\vspace{-1mm}
\section{Experiments}\label{results}
\vspace{-1mm}
In our experiments, we use a single VQ-VAE model designed using convolutional neural networks to achieve a compression factor of 16 ($f=4$), with a fixed codebook size of 8192, with pretrained weights from the work~\cite{paella}. Note that both the context and image are represented using the same codebook and VQ-VAE. 
We extend our experiments on different datasets, including Cityscapes~\cite{cityscapes}, Flickr~\cite{Flickr}, and STL-10~\cite{STL10} datasets. For comparison, we test our performance against conventional image compression techniques such as JPEG and JPEG 2000, and existing deep learning based semantic communication models such as Diff-GO+~\cite{diffgo+}, GESCO~\cite{gesco}, Diff-GO~\cite{diffgo}, OD (Original Diffusion)~\cite{diffgo}, and LaMI-GO~\cite{lamigo}. We also compare with semantic image syntheses models including SPADE~\cite{SPADE}, CC-FPSE~\cite{CC-FPSE}, SMIS~\cite{SMIS}, OASIS~\cite{OASIS}, and SDM~\cite{SDM}. 

We conduct the experiments in the follow scenarios:
\begin{itemize}
    \item \textbf{Scenario-1}: We evaluate our model's performance on a set of predefined tasks: image reconstruction, 
    image classification, and object detection. For image reconstruction,
    the Cityscapes dataset is used with an input resolution of $3\times256\times512$. Image classification is implemented in the STL-10 dataset with a resolution of $3\times224\times224$. For the classification, a ResNet-18 architecture is adopted. For object detection, the Flickr dataset with an image resolution of $3\times224\times224$ is considered. We adopt Fast R-CNN for object detection.
    \item \textbf{Scenario-2}: In this setup, we analyze the effect of our local semantic feedback (LSF) on Scenario-1. Specifically, we re-evaluate the performance of image classification and object detection tasks after implementing the LSF.
    \item \textbf{Scenario-3}: In this setup, we assume that the transmitter and receiver pair are optimized for classification in the STL-10 data set. The receiver is then interested in object detection on the same dataset. This experiment evaluates our model's ability to adapt to new tasks.
\end{itemize}
\vspace{-2mm}
\subsection{Scenario-1}
\subsubsection{Image Reconstruction }
The overall performance of image reconstruction is illustrated in
Table~\ref{tab:tablerec} where perceptual metrics are considered. 
The NA entry is used to represent non-diffusion-based approaches that use single inference steps. The models evaluated under the same conditions are represented by {\textdagger}. From the results, our proposed framework outperform all the other complex models in both the reconstruction quality and the number of steps for data regeneration. We also evaluate the bandwidth cost for image reconstruction in Table. \ref{tab:tableBW}, where our proposed method achieves superior performance with the best bandwidth efficiency.

\begin{table}[!t]
\caption{Measured semantic similarity of the reconstructed images for the Cityscapes. Part of the results is from the work~\cite{lamigo}} 
\label{tab:tablerec}
\centering
\begin{tabular}{|c||c|c|c|c|}
\hline
Method & LPIPS$\downarrow$ & FID$\downarrow$ & Steps$\downarrow$   \\
\hline
SPADE~\cite{SPADE}& 0.546 & 103.24 & NA\\
\hline
CC-FPSE~\cite{CC-FPSE}& 0.546 & 245.9& NA\\
\hline
SMIS~\cite{SMIS}& 0.546 & 87.58& NA\\
\hline
OASIS~\cite{OASIS}& 0.561 & 104.03& NA\\
\hline
SDM~\cite{SDM}& 0.549 & 98.99& NA\\
\hline
OD\textdagger~\cite{diffgo}& 0.2191 & 55.85& 1000\\
\hline
GESCO\cite{gesco}\textdagger & 0.591 & 83.74& 1000\\
\hline
RN\textdagger \cite{diffgo}&  0.3448 & 96.409& 1000\\
\hline
Diff-GO\textdagger (n=20)\cite{diffgo}& 0.3206 & 74.09& 1000\\
\hline
Diff-GO\textdagger (n=50)\cite{diffgo}& 0.2697 & 72.95& 1000\\
\hline
Diff-GO\textdagger (n=100)\cite{diffgo}& 0.2450 & 68.59& 1000\\
\hline
Diff-GO+\textdagger (W = 64, L = 32)\cite{diffgo+} & 0.2231 & 60.93& 1000 \\
\hline
Diff-GO+\textdagger (W = 128, L = 32)\cite{diffgo+} & 0.2126 & 58.20 & 1000\\
\hline
TACO (Ours)
& \textbf{0.056} & \textbf{11.21} & NA\\
\hline
\end{tabular}
\vspace{-3mm}
\end{table}
\begin{table}[!t]
\caption{Bandwidth comparison in Cityscape dataset.  } 
\label{tab:tableBW}
\centering
\begin{tabular}{|c||c|c|c|}
\hline
\textbf{Method}& \textbf{Bandwidth} (KB) & \textbf{FID}$\downarrow$     \\
\hline
Original image& 187.63  & - \\
\hline
GESCO& 14.52 & 83.75\\
\hline
Diff-GO+(n=1024,L=1024)& 13.5 & 68.25\\

+GO-EVAE(n=1024,L=16)&  &  \\
\hline
OD & 1994.41 & 55.85\\
\hline
Diff-GO+(n=128,L=32)& 87.04 & 58.20\\
\hline
\textbf{TACO} & \textbf{11.70} & \textbf{34.30}  \\

\hline
\end{tabular}
\vspace{-6mm}
\end{table}

\subsubsection{Classification}
We now present the performance of classification as a downstream task in Table~\ref{tab:acc}. 
From the results, our proposed method perform better than other approaches under similar bandwidth. Moreover,
as we add more task-specific information, the classification accuracy further increases. 
Particularly, we see a significant improvement using our proposed method at the very low level of bandwidth usage.
\begin{table}[t]
\caption{Classification accuracy on STL-10 dataest. }
\vspace{-5mm}
\begin{center}
\begin{tabular}{|c|c|c|}
\hline
\textbf{Model} & \textbf{Accuracy $(\%)$} &\textbf{Bandwidth (KB)}\\
\hline
Original image & 94.30 & 59.93 \\
\hline
TACO w/o LSF ($\zeta$) & 80.31 & 0.311 \\
\hline

TACO w/o LSF ($\zeta$ + 10\% info. ) & 85.51 & 1.51 \\
\hline
TACO w/o LSF ($\zeta$ + 20\% info. ) & 89.28 & 2.24 \\
\hline
TACO w/o LSF ($\zeta$ + 30\% info. ) & 91.19 & 3.10 \\
\hline
\textbf{TACO-LSF} & \textbf{96.49} & \textbf{0.65} \\
\hline
JPEG(q=1) & 16.51 & 1.59 \\
\hline
JPEG(q=5) & 22.82 & 1.89 \\
\hline
JPEG(q=20) & 89.09 & 3.28\\
\hline
JPEG(q=50) & 92.64 & 4.56 \\
\hline
JPEG2000(q=100) & 70.25 & 1.47 \\
\hline
JPEG2000(q=60) & 81.97 & 2.45 \\
\hline
JPEG2000(q=40) & 88.22 & 3.66\\
\hline
JPEG2000(q=30) & 90.59 & 4.88 \\

\hline
\end{tabular}
\label{tab:acc}
\end{center}
\vspace{-5mm}
\end{table}

\vspace{-4mm}
\subsubsection{Object detection}
In Table~\ref{tab:object}, we present the performance of object detection compared to classical image compression techniques.  From the table, we see that our proposed model performs better in all three metrics.

\begin{table}[t]
\caption{Object detection on Flickr dataset.}
\vspace{-5mm}
\begin{center}
\begin{tabular}{|c|c|c|c|}
\hline
\textbf{Model}  & \textbf{mIoU}&\textbf{mAP}&  \textbf{Bandwidth (KB)}\\
\hline

TACO w/o LSF ( $\zeta$) & 0.12&0.09 & 0.311 \\
\hline

TACO w/o LSF ($\zeta$ + 10\% info. ) & 0.48 & 0.39 & 1.49 \\
\hline
TACO w/o LSF ($\zeta$ + 20\% info. ) & 0.62 & 0.52 & 2.54 \\
\hline
TACO w/o LSF ($\zeta$ + 50\% info. ) & 0.74 & 0.64 & 5.52 \\
\hline
TACO w/o LSF ($\zeta$ + 70\% info. ) & 0.77 & 0.69 & 7.38 \\
\hline
\textbf{TACO-LSF} & \textbf{0.78} & \textbf{0.80} & \textbf{5.93} \\
\hline
JPEG(q=1) & 0.0 & 0.0 & 1.80 \\
\hline
JPEG(q=10) & 0.1 & 0.06 & 3.11  \\
\hline
JPEG(q=20) & 0.43 & 0.33& 4.55 \\
\hline
JPEG(q=50) & 0.73 & 0.64& 7.65  \\
\hline
JPEG2000(q=100) & 0.10 & 0.07 & 1.59 \\
\hline
JPEG2000(q=80) & 0.16 & 0.11 & 1.98\\
\hline
JPEG2000(q=60) & 0.28 & 0.20&2.63 \\
\hline
JPEG2000(q=40) & 0.43 & 0.33 &3.95\\
\hline
JPEG2000(q=20) & 0.64 & 0.54 & 7.9 \\
\hline
\end{tabular}
\label{tab:object}
\end{center}
\vspace{-8mm}
\end{table}


\subsection{Scenario-2: Effect of the Local Semantic Feedback (LSF)}
Next, we investigate the effect of integrating the LSF at the transmitter. We empirically evaluate the performance and bandwidth improvement with LSF. The results of applying LSF for classification and object detection are reported in Table~\ref{tab:acc} and Table~\ref{tab:object}, respectively. From the results, we see a substantial improvement in both classification accuracy and bandwidth saving, which indicates information redundancy for 
classification and object detection. Note that our LSF shows better performance than the original images. It also shows that LSF helps emphasize the regions where the attention is mostly needed and blurs out the confusing background.





\begin{table}[t]
\caption{Object detection in Scenario-3.}
\vspace{-5mm}
\begin{center}
\begin{tabular}{|c|c|c|c|}
\hline
\textbf{Model} & \textbf{mIoU $(\%)$} & \textbf{mAP $(\%)$}&\textbf{Bandwidth (KB)}\\
\hline
Original image &- & -& 59.93 \\
\hline
Baseline optimized &23.1 & 24.0&0.65 \\
for classification & & &\\
\hline
JPEG(q=5) & 3.0&3.0& 1.89\\
\hline
JPEG2000(q=100) & 18.0&18.0& 1.47\\
\hline
\textbf{TACO} \textbf{(Feedback )}& \textbf{50.49} & \textbf{48.04} & \textbf{1.71} \\
\hline
\end{tabular}
\label{tab:cl_fb}
\end{center}
\vspace{-5mm}
\end{table}

\subsection{Scenario-3: New Task at the Receiver}
\begin{figure}[t]

    \centering
    \subfigure[Original image.]{\includegraphics[width=0.13\textwidth]{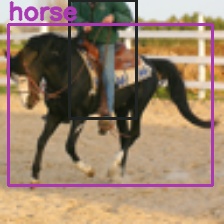}}
    \hfill
    \subfigure[Context image.]{\includegraphics[width=0.13\textwidth]{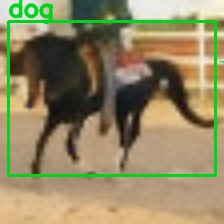}}
    \hfill
    \subfigure[After feedback.]{\includegraphics[width=0.13\textwidth]{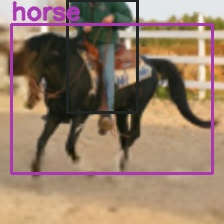}}
    \caption{Object Detection for Task Switching in Different Steps}
    \label{fig:three_images}
    \vspace{-5mm}
\end{figure}

In this part, we present the results in the scenario of task change. Specifically, we consider that the communication system is originally optimized for classification and switch interest to object detection. 
Fig.~\ref{fig:three_images} represents a scenario where Fig.~\ref{fig:three_images}-(a) represents the object detection on the actual image. At the receiver, object detection on the context image outputs a low confidence in Fig.~\ref{fig:three_images}-(b). As a result, the receiver requests more resolution in the selected box region. Fig.~\ref{fig:three_images}-(c) represents the object detection results after one round of communication. Note that the receiver transmits 4 integer numbers and the transmitter's response is in the latent domain, which is bandwidth-efficient. Table~\ref{tab:cl_fb} illustrates performance after the above setups. 
From the results, the proposed method with transmitter feedback outperforms other methods.


\vspace{-3mm}
\subsection{Evaluation of Reconstruction Time}
Table~\ref{tab:tabletime} summarizes the reconstruction time on average image generation for the Cityscapes dataset with a resolution of $3\times256\times512$ on a single NVIDIA A100 GPU. From the table, we see a substantial performance improvement compared to the diffusion-based models. The main advantage of our method is that a simple architecture while providing flexibility in conveying context with task-specific information.

\begin{table}[!t]
\caption{Complexity of Image Reconstruction} 
\label{tab:tabletime}
\centering
\begin{tabular}{|c||c|c|c|c|c|c|}
\hline
Model & Steps (forward/backward)& FID$\downarrow$  & LPIPS$\downarrow$& Time (s)$\downarrow$\\
\hline
Diff-GO+& 30/30 &65.81&0.3546&7.2 \\
\hline
LaMI-GO& 4/8 &29.84&0.1327&1.29\\ 
\hline
Ours& - &11.21&0.056&0.0015\\ 
\hline

\end{tabular}
\vspace{-5mm}
\end{table}

\section{Conclusion and Future works}
In this work, we introduce a novel semantic communication framework with task adaption and context embedding, which achieves a superior performance in both predefined downstream tasks and unknown new tasks. We introduce a new interpretation of the problem formulation for semantic communication and present a generic framework via a light-weighted generative model, which drastically reduces the information recovery latency. In future work, we shall extend our formulation with channel noise and analyze its effects in our system. Another promising direction is to explore the compression of the context representation and investigate different transformations, such as the discrete wavelet transformation.

\bibliographystyle{IEEEtran}
\bibliography{IEEEexample}

\begin{thebibliography}{10}
\providecommand{\url}[1]{#1}
\csname url@samestyle\endcsname
\providecommand{\newblock}{\relax}
\providecommand{\bibinfo}[2]{#2}
\providecommand{\BIBentrySTDinterwordspacing}{\spaceskip=0pt\relax}
\providecommand{\BIBentryALTinterwordstretchfactor}{4}
\providecommand{\BIBentryALTinterwordspacing}{\spaceskip=\fontdimen2\font plus
\BIBentryALTinterwordstretchfactor\fontdimen3\font minus \fontdimen4\font\relax}
\providecommand{\BIBforeignlanguage}[2]{{%
\expandafter\ifx\csname l@#1\endcsname\relax
\typeout{** WARNING: IEEEtran.bst: No hyphenation pattern has been}%
\typeout{** loaded for the language `#1'. Using the pattern for}%
\typeout{** the default language instead.}%
\else
\language=\csname l@#1\endcsname
\fi
#2}}
\providecommand{\BIBdecl}{\relax}
\BIBdecl

\bibitem{mm1}
A.~Br{\"o}ring, V.~Kulkarni, A.~Zirkler, P.~Buschmann, K.~Fysarakis, S.~Mayer, B.~Soret, L.~D. Nguyen, P.~Popovski, S.~Samarakoon \emph{et~al.}, ``Intelliot: intelligent iot environments,'' in \emph{Global IoT Summit}, Dublin, Ireland, 2022, pp. 55--68.

\bibitem{diffgo}
A.~Wijesinghe, S.~Zhang, S.~Wanninayaka, W.~Wang, and Z.~Ding, ``Diff-go: Diffusion goal-oriented communications with ultra-high spectrum efficiency,'' in \emph{2024 IEEE International Conference on Communications Workshops (ICC Workshops)}, Denver, CO, USA, 2024, pp. 1079--1084.

\bibitem{9398576}
H.~Xie, Z.~Qin, G.~Y. Li, and B.-H. Juang, ``Deep learning enabled semantic communication systems,'' \emph{IEEE Transactions on Signal Processing}, vol.~69, pp. 2663--2675, 2021.

\bibitem{9322296}
------, ``Deep learning based semantic communications: An initial investigation,'' in \emph{GLOBECOM 2020}, Taipei, Taiwan, 2020, pp. 1--6.

\bibitem{10038754}
Z.~Weng, Z.~Qin, X.~Tao, C.~Pan, G.~Liu, and G.~Y. Li, ``Deep learning enabled semantic communications with speech recognition and synthesis,'' \emph{IEEE Transactions on Wireless Communications}, vol.~22, no.~9, pp. 6227--6240, 2023.

\bibitem{gesco}
E.~Grassucci, S.~Barbarossa, and D.~Comminiello, ``Generative semantic communication: Diffusion models beyond bit recovery,'' \emph{arXiv preprint arXiv:2306.04321}, 2023.

\bibitem{diffgon}
S.~Wanninayaka, A.~Wijesinghe, W.~Wang, Y.-C. Chao, S.~Zhang, and Z.~Ding, ``Diff-go\textsuperscript{n}: Enhancing diffusion models for goal-oriented communications,'' \emph{arXiv preprint arXiv:2412.06980}, 2024.

\bibitem{diffgo+}
A.~Wijesinghe, S.~Zhang, S.~Wanninayaka, W.~Wang, and Z.~Ding, ``Diff-go+: An efficient diffusion goal-oriented communication system with local feedback,'' \emph{IEEE Transactions on Wireless Communications}, 2025.

\bibitem{vqvae}
A.~Van Den~Oord, O.~Vinyals \emph{et~al.}, ``Neural discrete representation learning,'' in \emph{Advances in Neural Information Processing Systems}, vol.~30, Long Beach, CA, USA, 2017.

\bibitem{lamigo}
A.~Wijesinghe, S.~Wanninayaka, W.~Wang, Y.-C. Chao, S.~Zhang, and Z.~Ding, ``Lami-go: Latent mixture integration for goal-oriented communications achieving high spectrum efficiency,'' \emph{arXiv preprint arXiv:2412.17839}, 2024.

\bibitem{mahadi1}
L.~Qiao, M.~B. Mashhadi, Z.~Gao, R.~Tafazolli, M.~Bennis, and D.~Niyato, ``Token communications: A unified framework for cross-modal context-aware semantic communications,'' \emph{arXiv preprint arXiv:2502.12096}, 2025.

\bibitem{mahadi2}
L.~Qiao, M.~B. Mashhadi, Z.~Gao, and D.~G{\"u}nd{\"u}z, ``Token-domain multiple access: Exploiting semantic orthogonality for collision mitigation,'' \emph{arXiv preprint arXiv:2502.06118}, 2025.

\bibitem{multi1}
Y.~Sheng, F.~Li, L.~Liang, and S.~Jin, ``A multi-task semantic communication system for natural language processing,'' in \emph{2022 IEEE 96th Vehicular Technology Conference}, London/Beijing, 2022, pp. 1--5.

\bibitem{gradcam}
R.~R. Selvaraju, M.~Cogswell, A.~Das, R.~Vedantam, D.~Parikh, and D.~Batra, ``Grad-cam: Visual explanations from deep networks via gradient-based localization,'' in \emph{Proceedings of the IEEE International Conference on Computer Vision}, Venice, Italy, 2017, pp. 618--626.

\bibitem{lpips}
R.~Zhang, P.~Isola, A.~A. Efros, E.~Shechtman, and O.~Wang, ``The unreasonable effectiveness of deep features as a perceptual metric,'' in \emph{Proceedings of the IEEE Conference on Computer Vision and Pattern Recognition}, Salt Lake City, Utah, USA, 2018, pp. 586--595.

\bibitem{fid}
M.~Heusel, H.~Ramsauer, T.~Unterthiner, B.~Nessler, and S.~Hochreiter, ``Gans trained by a two time-scale update rule converge to a local nash equilibrium,'' in \emph{Advances in Neural Information Processing Systems}, vol.~30, Long Beach, CA, USA, 2017.

\bibitem{paella}
D.~Rampas, P.~Pernias, and M.~Aubreville, ``A novel sampling scheme for text-and image-conditional image synthesis in quantized latent spaces,'' \emph{arXiv preprint arXiv:2211.07292}, 2022.

\bibitem{cityscapes}
M.~Cordts, M.~Omran, S.~Ramos, T.~Rehfeld, M.~Enzweiler, R.~Benenson, U.~Franke, S.~Roth, and B.~Schiele, ``The cityscapes dataset for semantic urban scene understanding,'' in \emph{Proceedings of the IEEE CVPR}, Las Vegas, NV, USA, 2016, pp. 3213--3223.

\bibitem{Flickr}
P.~Young, A.~Lai, M.~Hodosh, and J.~Hockenmaier, ``From image descriptions to visual denotations: New similarity metrics for semantic inference over event descriptions,'' \emph{Transactions of the Association for Computational Linguistics}, vol.~2, pp. 67--78, 2014.

\bibitem{STL10}
A.~Coates, A.~Ng, and H.~Lee, ``An analysis of single-layer networks in unsupervised feature learning,'' in \emph{Proceedings of the Fourteenth International Conference on Artificial Intelligence and Statistics}, Fort Lauderdale, FL, USA, 2011, pp. 215--223.

\bibitem{SPADE}
T.~Park, M.-Y. Liu, T.-C. Wang, and J.-Y. Zhu, ``Semantic image synthesis with spatially-adaptive normalization,'' in \emph{2019 IEEE/CVF CVPR}, Long Beach, CA, USA, 2019, pp. 2332--2341.

\bibitem{CC-FPSE}
X.~Liu, G.~Yin, J.~Shao, X.~Wang \emph{et~al.}, ``Learning to predict layout-to-image conditional convolutions for semantic image synthesis,'' in \emph{NeurIPS}, vol.~32, Vancouver, Canada, 2019.

\bibitem{SMIS}
Z.~Zhu, Z.~Xu, A.~You, and X.~Bai, ``Semantically multi-modal image synthesis,'' in \emph{Proceedings of the IEEE/CVF Conference on Computer Vision and Pattern Recognition}, 2020, pp. 5467--5476.

\bibitem{OASIS}
V.~Sushko, E.~Sch{\"o}nfeld, D.~Zhang, J.~Gall, B.~Schiele, and A.~Khoreva, ``You only need adversarial supervision for semantic image synthesis,'' \emph{arXiv preprint arXiv:2012.04781}, 2020.

\bibitem{SDM}
W.~Wang, J.~Bao, W.~Zhou, D.~Chen, D.~Chen, L.~Yuan, and H.~Li, ``Semantic image synthesis via diffusion models,'' \emph{arXiv preprint arXiv:2207.00050}, 2022.

\end{thebibliography}

\end{document}